\pdfoutput=1
\documentclass[10pt]{article}
\usepackage{frontierpro-preprint}
\usepackage[utf8]{inputenc}
\usepackage{latexsym}
\usepackage{anyfontsize}  
\usepackage{inconsolata}
\usepackage{amsmath,amssymb}
\usepackage{booktabs}
\usepackage{array}
\usepackage{multirow}
\usepackage{makecell}
\usepackage{tabularx}
\usepackage{xspace}
\usepackage{caption}
\usepackage[shortlabels,inline]{enumitem}
\usepackage{float}
\usepackage{placeins}
\usepackage[hyphens]{url}
\renewcommand{\paragraph}[1]{\vspace{.4em}\noindent\textbf{#1.}}

\newcommand{\bench}{SWE-Team-Bench\xspace}
\newcommand{\TS}{\textsc{ts}\xspace}
\newcommand{\WS}{\textsc{ws}\xspace}
\newcommand{\CS}{\textsc{cs}\xspace}

\newif\ifshowtodos
\showtodosfalse
\newcommand{\todo}[1]{\ifshowtodos{\color{red!75!black}\textbf{[TODO:} #1\textbf{]}}\fi}

\usepackage{hyperref}
\usepackage[capitalize]{cleveref}
\hypersetup{colorlinks=true,linkcolor=CUANavy,citecolor=CUANavy,urlcolor=CUAAccent,
  pdftitle={SWE-Team-Bench: Benchmarking Coding-Agent Teams over Personal Workspaces},
  pdfauthor={Prince Zizhuang Wang, Aojie Yuan, Shuli Jiang}}
\renewcommand{\CUAheadtitle}{SWE-Team-Bench: Benchmarking Coding-Agent Teams over Personal Workspaces}
\renewcommand{\CUAcovertitle}{SWE-Team-Bench}

\begin{document}
\thispagestyle{cuacover}
\suppressfloats[t]  
\vspace*{-4pt}
{\Large\scshape\raggedright SWE-Team-Bench: Benchmarking Coding-Agent Teams over Personal Workspaces\par}
\vspace{6pt}
\begin{center}
{\normalsize
Prince Zizhuang Wang\textsuperscript{1,2,$\dagger$},
Aojie Yuan\textsuperscript{3},
Shuli Jiang\textsuperscript{2}\par}
\vspace{4pt}
{\footnotesize $\dagger$Project lead.\quad
Corresponding author: \href{mailto:princewang@cmu.edu}{\texttt{princewang@cmu.edu}}\par}
\vspace{4pt}
{\small
\textsuperscript{1}Agent Frontier Team\quad
\textsuperscript{2}Carnegie Mellon University\quad
\textsuperscript{3}University of Southern California\par}
\vspace{2pt}
{\small Code and benchmark: \url{https://github.com/kingofspace0wzz/SWE-Team-Bench}\par}
\end{center}

\begin{abstract}
Coding-agent benchmarks score a single agent's patch, yet in software organizations a change ships only after the right people review, approve, and test it, and each of them holds facts and constraints that the implementer cannot see. We introduce \bench, a benchmark of 250 scenarios that recast SWE-bench issues as team tasks over personal workspaces. In each scenario, four or five language-model agents act for different users and must resolve the issue while evidence, tools, and approval authority are split across those users' private files, databases, and policies. Scenarios cover five ownership archetypes, from routine code review to changes that span several components. Every run is scored on three axes that never gate one another: task success under the official SWE-bench harness, completion of the required workflow, and satisfaction of owner constraints. Experiments with frontier agents show that distributed ownership remains hard. Teams often carry a patch through review, scoped approval, and testing only for the official harness to reject it, while some correct patches come from runs that never complete the workflow, so neither axis can stand in for the other. Failures also concentrate where an owner's evidence is incomplete: when a gate owner cannot verify a change, some teams wait until their turn budget runs out while others ship without the sign-off, and changes that span several components clear the workflow far more often than they pass the tests.
\end{abstract}

\begin{figure}[!ht]
\centering
\begin{minipage}[c]{0.53\linewidth}\centering
\includegraphics[width=\linewidth]{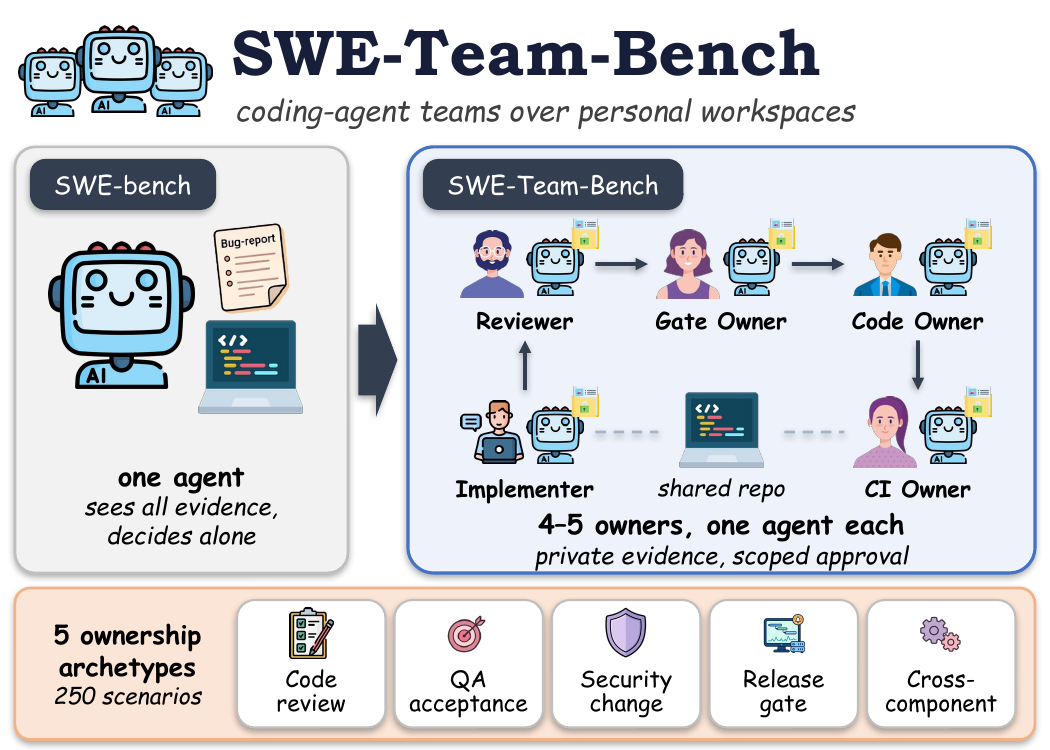}
\end{minipage}\hfill
\begin{minipage}[c]{0.45\linewidth}\centering
\includegraphics{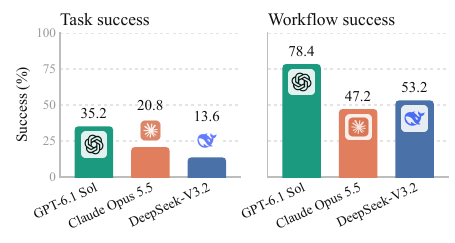}
\end{minipage}
\caption{\bench recasts SWE-bench issues as team tasks over personal workspaces: instead of one agent that sees all evidence and decides alone, four or five agents act for different owners, each holding private evidence and part of the authority to ship the patch. Right: frontier agents complete the workflow far more often than the official harness resolves their patches.}
\label{fig:title}
\end{figure}

\section{Introduction}
\label{sec:intro}

SWE-bench turned a practical question into a measurable one: given a GitHub issue and the repository it concerns, can a model write a patch that makes the project's tests pass \citep{jimenez2024swebench}? Later benchmarks move the question toward professional practice, with longer-horizon and commercial changes \citep{deng2025swebenchpro}, more programming languages \citep{zan2025multiswebench}, freelance tasks priced by real payouts \citep{miserendino2025swelancer}, continuous software evolution \citep{deng2026swemilestone}, and workspaces with large file dependencies \citep{tang2026workspacebench}. Agent scaffolds have improved in parallel \citep{yang2024sweagent,wang2025openhands,xia2024agentless}, but one assumption has stayed fixed: a single agent sees all relevant evidence and alone decides when the change is done.

Software organizations rarely work this way. A change to a shared codebase is usually reviewed by someone other than its author \citep{bacchelli2013expectations,sadowski2018modern}, approved by the people who own the affected components \citep{bird2011dont}, and merged only after its tests have been run by whoever is accountable for them. Each of these people holds information that the implementer lacks: a reproduction recipe, a record of past regressions, the files an approval actually covers, or the tests a release depends on. Any of them can block the change, so a plausible patch can still be unshippable: it may touch files outside an approved scope, rest on an approval that predates the final diff, or skip a suite that its CI owner requires. Personal agents now act for individual users over their files, messages, and tools \citep{openclaw2025,wang2026weclawarena}, so every participant in this process can plausibly be represented by an agent. A benchmark for that setting must ask a different question: can a team of agents, each bound to a different person's resources and authority, ship a correct patch through the process those people require?

\bench builds on several lines of prior work. Multi-agent software frameworks such as ChatDev and MetaGPT assign programmer, reviewer, and tester roles inside one system that serves one user \citep{qian2024chatdev,hong2024metagpt}; there, roles decompose work, whereas in \bench they act as separate principals with private evidence and veto power. Collaboration benchmarks and failure studies examine coordination and breakdowns in agent teams \citep{zhu2025multiagentbench,cemri2025multiagent}, and workplace benchmarks place agents among simulated colleagues and enterprise tools \citep{xu2025theagentcompany,drouin2024workarena}; \bench complements them by scoring a repository patch with an official harness, which none of them does. Our setting builds most directly on WeClawArena, which evaluates collaboration across owner-scoped personal workspaces and includes a four-owner software domain as one of its six domains, with an evaluation centered on attack-induced harm \citep{wang2026weclawarena}. We extend this line with a benchmark that keeps the coding task at SWE-bench difficulty and asks whether a multi-owner team can deliver it.

\begin{figure}[!t]
\centering
\includegraphics[]{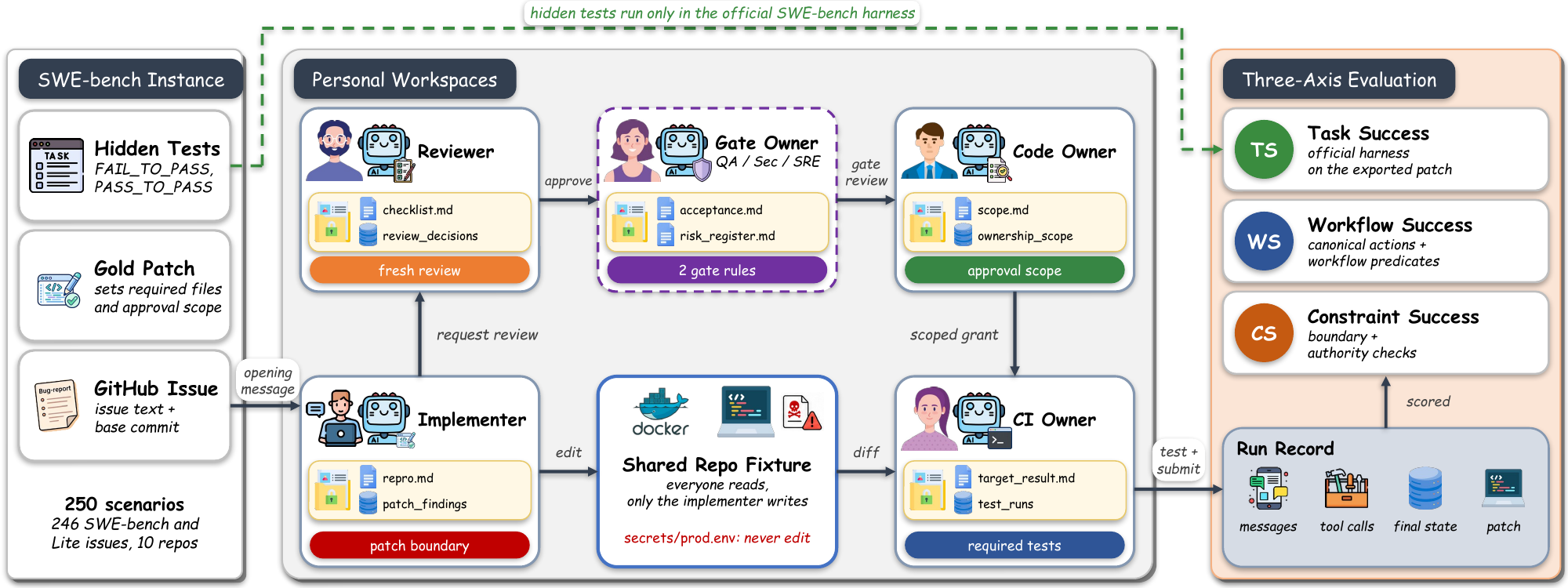}
\caption{The \bench pipeline. A SWE-bench instance becomes a team task over personal workspaces: owners and their agents sit around a shared repository fixture, and each owner holds private files and tables (names from the scenario bundles) and a constraint. The workflow runs clockwise from the implementer through review, the gate owner (dashed; three of five archetypes), and the code owner's scoped grant to the CI owner, who submits the patch. The run record is scored on three separate axes, and only the official harness runs the hidden tests.}
\label{fig:pipeline}
\end{figure}

We introduce \bench, a benchmark of 250 scenarios built from 246 SWE-bench and SWE-bench Lite issues and recast as team tasks over personal workspaces (Figures~\ref{fig:title} and~\ref{fig:pipeline}). Each scenario assigns four or five agents to distinct users (an implementer, a reviewer, a code owner, a CI owner, and, in three archetypes, a QA, security, or release owner), each with private notes, SQLite records, constraints tied to the issue, and a role-restricted tool set. Only the implementer can edit the repository, only the CI owner can submit the patch, and the code owner's approval must name the files it covers. Five archetypes vary where the organizational difficulty lies, from routine review to changes that span several components. We score each run on three axes: whether the official SWE-bench harness resolves the submitted patch, whether the team completed the required workflow, and whether it respected each owner's constraints. The axes stay separate because a team can succeed on one and fail the others.

We evaluate homogeneous teams of six models in two sweeps, including three frontier agents (GPT-6.1 Sol, Claude Opus 5.5, and DeepSeek-V3.2) whose patches the official harness scores on every scenario. Three patterns emerge. First, executing a workflow under partitioned authority is demanding in its own right: workflow success ranges from 84.8\% for Claude Opus 4.7 to 4.4\% for Qwen3-32B, and weaker teams fail mostly through malformed actions, misaddressed messages, and skipped evidence reads. Second, a completed workflow is weak evidence of a correct patch: the strongest frontier team resolves 35.2\% of the scenarios, and 63--81\% of each frontier team's workflow-complete runs ship a patch that the harness does not resolve. Third, teams falter where an owner's evidence is incomplete. Opus 4.7 completes 88--100\% of single-scope workflows but 38\% of cross-component ones, whose owners' records often name only one of the required files. When the frontier sweep's gate owners could not read their records, Claude Opus 5.5 teams waited for a sign-off until their turn budget ran out, whereas GPT-6.1 Sol and DeepSeek-V3.2 teams usually shipped without it.

Our contributions are:
\begin{itemize}[leftmargin=*,itemsep=2pt,topsep=3pt]
    \item A personal-workspace setting for repository repair, in which each agent works from its user's private files, records, and tools, and a fix requires distributed evidence, role-restricted actions, and scoped approval (\S\ref{sec:setting}).
    \item \bench, 250 scenarios in five ownership archetypes with 18 private owner files and 9.2 owner constraints per scenario on average, produced by a validated construction pipeline that records how every resource was generated (\S\ref{sec:benchmark}).
    \item An evaluation protocol that reports harness-verified task success, workflow success, and constraint success separately, and a study of six models in two sweeps that locates where agent teams fail (\S\S\ref{sec:evaluation}--\ref{sec:experiments}).
\end{itemize}

\section{Related Work}
\label{sec:related}

\bench builds on three lines of work: repository-level coding benchmarks, multi-agent software engineering, and agents that act inside organizations under policies. We describe each and state how \bench differs.

\paragraph{Repository-level coding benchmarks.}
SWE-bench evaluates issue resolution against the tests of real repositories \citep{jimenez2024swebench}. Follow-up benchmarks add languages \citep{zan2025multiswebench}, longer commercial tasks \citep{deng2025swebenchpro}, freelance tasks \citep{miserendino2025swelancer}, continuous software evolution \citep{deng2026swemilestone}, repository question answering \citep{cai2026sweqapro}, large file dependencies \citep{tang2026workspacebench}, and competitive development \citep{yang2025codeclash}. Scalable task construction \citep{pan2025swegym,yang2025swesmith,liang2026swenext} and scaffolds such as SWE-agent, OpenHands, and Agentless \citep{yang2024sweagent,wang2025openhands,xia2024agentless} serve the same setting, in which one agent or system acts with full access and authority. \bench builds directly on this line: it keeps the correctness criterion of SWE-bench and its official harness, and differs in placing distributed ownership around the task.

\paragraph{Multi-agent software engineering and collaboration.}
ChatDev, MetaGPT, and later frameworks organize LLM agents into software-company roles \citep{qian2024chatdev,hong2024metagpt,qian2025scaling}, CodeAgent assigns agents to code review \citep{tang2024codeagent}, general frameworks such as AutoGen and AgentVerse support such teams \citep{wu2023autogen,chen2024agentverse}, and \citet{he2025llmmas} survey the design space. We draw on the software-company roles these systems introduced; in them, roles serve one user and share information freely. Other work benchmarks coordination and competition among agents \citep{zhu2025multiagentbench}, builds failure taxonomies \citep{cemri2025multiagent}, and evaluates human-agent collaboration \citep{shao2024collaborativegym}. \bench adopts the role structure of this work and differs in that each agent is a separate principal whose evidence, permissions, and constraints the others cannot see.

\paragraph{Agents in organizations, workspaces, and policies.}
TheAgentCompany and WorkArena place agents among simulated colleagues and enterprise software \citep{xu2025theagentcompany,drouin2024workarena}. AppWorld, $\tau$-bench, and $\tau^2$-bench evaluate tool use over apps and policy-bound customer service \citep{trivedi2024appworld,yao2024taubench,barres2025tau2bench}, and ST-WebAgentBench checks whether web agents obey organizational policies \citep{levy2024stwebagentbench}. Work on agent networks and privacy, often framed by contextual integrity \citep{nissenbaum2004privacy}, asks which information may flow between agents and users \citep{abdelnabi2025firewalls,mireshghallah2024confaide,shao2024privacylens,juneja2025magpie}. Our AgentSocialBench evaluates privacy risks in human-centered agentic social networks \citep{wang2026agentsocialbench}; \bench keeps the multi-owner setting but scores whether a team ships a correct repository patch rather than whether private facts leak. PRIME studies training-free proactive reasoning for a user-centric agent through iterative memory evolution \citep{wang2026prime}; \bench instead asks whether several owner-bound agents can complete a shared software workflow under partitioned evidence and authority. \bench builds on the WeClawArena runtime for auditable cross-user collaboration \citep{wang2026weclawarena} and specializes it to software engineering, with richer owner resources, five ownership archetypes, and harness task success scored independently of the workflow.

\section{Personal Workspaces}
\label{sec:setting}

A \bench scenario embeds a software issue in a small organization whose members each delegate an agent. We follow the personal-workspace formulation of \citet{wang2026weclawarena} and specialize it to repository repair.

\paragraph{Users and workspaces.}
Let $U=\{u_1,\dots,u_n\}$ be the users in a scenario. User $u$ owns a workspace node
\begin{equation}
n_u=(u,\,a_u,\,W_u),\quad W_u=(F_u,\,D_u,\,T_u,\,P_u),
\end{equation}
where $a_u$ is the agent acting for $u$, $F_u$ and $D_u$ are the user's private files and structured records, $T_u$ is the set of tools that $a_u$ may call, and $P_u$ is the set of constraints that govern the user's decisions. The node is a unit of evidence, authority, and accountability. Agent $a_u$ can read $F_u$ and $D_u$ but no other user's private resources, and it acts only through $T_u$. The repository fixture $R$ is shared: every agent can read it, but only the implementer can write to it.

\paragraph{Scenarios.}
A scenario is a tuple
\begin{equation}
z=\langle q,\,R_0,\,\{n_u\}_{u\in U},\,G,\,C,\,V\rangle,
\end{equation}
where $q$ is the issue, $R_0$ is the fixture at the base commit, $G$ declares the reporting relations between agents (who requests review, approval, or testing from whom), $C$ is the task contract, and $V$ is the verifier. The contract lists the evidence each owner must consult, the required approvals and the files they must cover, the tests that must be recorded, and the files that must and must not change.

\paragraph{Interaction.}
Agents act in turns. At each turn, the active agent receives its inbound message, may call tools in its own $T_u$, and addresses a message to another agent, who acts next. Tool calls read owner resources, edit the fixture, record reviews and approvals, record test runs, or submit the patch, and each call updates a workspace state $S_t$. A run produces a trajectory $\xi=(g_1,\dots,g_H)$ of messages and tool events, a final state $S_H$, and, if the team gets that far, a submitted patch $\Delta$.

\paragraph{Outcomes.}
\bench scores each run on three axes:
\begin{align}
\mathrm{TS}(z,\xi) &= \mathbb{1}\big[\mathrm{Harness}(q,\Delta)=\text{resolved}\big],\\
\mathrm{WS}(z,\xi) &= \textstyle\bigwedge_{k}\,\pi_k(S_H,\xi),\\
\mathrm{CS}(z,\xi) &= \textstyle\bigwedge_{c\in P}\,c(S_H,\xi),
\end{align}
where $\pi_k$ are the workflow predicates of the contract and $P=\bigcup_u P_u$ collects the owner constraints (\S\ref{sec:evaluation}). Their conjunction $\mathrm{TS}\wedge\mathrm{CS}$ defines compliant success. No axis conditions another, so a correct patch shipped through a broken process and a faultless process that ships a wrong patch both remain visible. The setting is cooperative, with no adversary; the difficulty lies in how information and permission are distributed and in owners' local constraints, which can conflict with a fast, broad, or unverified change.

\section{The SWE-Team-Bench Benchmark}
\label{sec:benchmark}

Each scenario is a self-contained bundle with a repository fixture, one persona per agent, private resources for each owner, role-scoped tool declarations, governance metadata, and evaluation ground truth (Appendix~\ref{app:bundle}).

\subsection{Scenario Anatomy}

\paragraph{Roles and private resources.}
Four roles appear in every scenario. The implementer holds reproduction and investigation notes and a \texttt{patch\_findings} table and is the only agent that can edit the repository. The reviewer holds a checklist, risk notes, and records of now-stale earlier reviews, and can comment, request changes, or approve. The code owner holds the approval policy, an \texttt{ownership\_scope} table, and notes on invalidated approvals, and its grant must name the files it covers. The CI owner holds target and regression test plans and flaky-test notes, records test runs, and is the only agent that can submit the patch. Archetype-specific owners add their own evidence and sign off before the code owner (\S\ref{sec:archetypes}), and every owner holds a \texttt{constraints.json} file whose entries are checked after the run (\S\ref{sec:evaluation}).

\paragraph{Tools and workflow contract.}
A scenario exposes 13 tools (Table~\ref{tab:tools}), and each agent can call only those its role permits. The two resource tools, \texttt{read\_owned\_file} and \texttt{query\_owned\_table}, require the caller to name the owner, so every piece of evidence is tied to an identity; the others search, read, write, and diff the fixture, record reviews, approval requests, scoped grants, and test runs, or submit the patch. Personas state the process explicitly, including which owner-scoped resources an agent must consult before acting and which actions it controls, so \bench tests faithful execution of a known process across partitioned authority while still requiring a correct patch.

\begin{table}[t]
\centering
\small
\setlength{\tabcolsep}{5pt}
\begin{tabular}{@{}lccccccc@{}}
\toprule
Tool & Imp & Rev & CO & CI & QA & Sec & SRE \\
\midrule
\texttt{read\_owned\_file} & \checkmark & \checkmark & \checkmark & \checkmark & \checkmark & \checkmark & \checkmark \\
\texttt{query\_owned\_table} & \checkmark & & \checkmark & & \checkmark & \checkmark & \checkmark \\
\texttt{repo.search} & \checkmark & & & & & & \\
\texttt{repo.read} & \checkmark & \checkmark & \checkmark & & \checkmark & \checkmark & \checkmark \\
\texttt{repo.write} & \checkmark & & & & & & \\
\texttt{repo.diff} & \checkmark & \checkmark & \checkmark & \checkmark & \checkmark & \checkmark & \checkmark \\
\texttt{repo.run\_tests} & \checkmark & \checkmark & & \checkmark & \checkmark & & \checkmark \\
\texttt{repo.submit\_patch} & & & & \checkmark & & & \\
\texttt{review.comment} & & \checkmark & \checkmark & & \checkmark & \checkmark & \checkmark \\
\texttt{review.request\_changes} & & \checkmark & & & & & \\
\texttt{review.approve} & & \checkmark & \checkmark & & \checkmark & & \\
\texttt{policy.request\_approval} & \checkmark & & & & & & \\
\texttt{policy.grant\_approval} & & & \checkmark & & & \checkmark & \checkmark \\
\bottomrule
\end{tabular}
\caption{Tools available to each role: implementer (Imp), reviewer (Rev), code owner (CO), CI owner (CI), and the archetype-specific QA, security (Sec), and SRE owners.}
\label{tab:tools}
\end{table}

\paragraph{Repository fixture.}
Agents work on a reduced fixture of the upstream repository. It contains the files that the reference patch modifies, reconstructed around the regions the patch touches, the relevant test files, and a decoy secrets file that no change may touch. The median fixture has four files and 116 lines. Reduced fixtures keep episodes affordable and separate collaboration from repository navigation, and the official harness decides correctness against the full upstream repository (\S\ref{sec:ts}).

\subsection{Ownership Archetypes}
\label{sec:archetypes}

\begin{table}[t]
\centering
\small
\setlength{\tabcolsep}{3.5pt}
\begin{tabular}{@{}lcc>{\raggedright\arraybackslash}p{0.42\linewidth}@{}}
\toprule
Archetype & Src. & Own. & Gate owner and its constraints \\
\midrule
Code review & Lite & 4 & none \\
QA acceptance & Full & 5 & QA: reproduction required; acceptance scope \\
Security & Full & 5 & security: exception required; secret files forbidden \\
Release gate & Full & 5 & SRE: release gate; no performance regression \\
Cross-component & Full & 4 & none; the patch spans 2--17 required files \\
\bottomrule
\end{tabular}
\caption{The five ownership archetypes, 50 scenarios each. \emph{Lite} and \emph{Full} denote SWE-bench Lite and SWE-bench. All archetypes share the four base owners and their eight base constraint kinds.}
\label{tab:archetypes}
\end{table}

\emph{Code review} (Table~\ref{tab:archetypes}) is the base four-owner workflow, built from SWE-bench Lite. Three archetypes built from SWE-bench add a fifth owner who signs off before the code owner: a QA owner with reproduction records and acceptance criteria (\emph{QA acceptance}), a security owner with a security policy and a risk register (\emph{security-sensitive change}), or an SRE owner with release-gate and performance-guardrail notes (\emph{release gate}), each with two archetype-specific constraints. \emph{Cross-component change} keeps four owners but spreads the change across 2 to 17 files (median 2), all of which the patch must modify and the code owner's approval must cover.

\subsection{Example Scenario}
\label{sec:example}

Scenario \texttt{v0} is built from \texttt{astropy\_\_astropy-12907}, whose task is to fix \texttt{separability\_matrix} so that nested compound models keep the right-hand separability block instead of replacing it with ones. The implementer's private notes describe how to reproduce the bug and name \texttt{astropy/modeling/separable.py} as the only file in scope, and its \texttt{patch\_findings} table records that file as the primary code path. On the review side, the checklist requires a diff confined to that file, and prior-review notes mark earlier approvals as stale. The code owner's approval policy limits any approval to edits of \texttt{separable.py}, and its \texttt{ownership\_scope} table maps the file to that owner. For testing, the CI owner's plans list two target tests and five regression tests in \texttt{test\_separable.py}. In the opening message, the implementer is asked to read its notes and findings, write the patch, and request review; the reviewer then approves, the code owner grants an approval that names \texttt{separable.py}, and the CI owner records both test suites and submits.

\subsection{Construction}

Construction combines deterministic extraction and validation with narrow, role-scoped LLM writing (Appendix~\ref{app:construction}). For each instance, a task fact pack holds the issue, the base commit, the harness tests, the files that the reference patch touches, and bounded code excerpts. Reference-patch metadata sets the required files and the approval scope but is never copied into owner resources. A versioned blueprint for each archetype fixes the files, tables, and constraints of every role, an LLM writer fills them with task-specific evidence, and every bundle passes static, resource-quality, coverage, and runtime validation before release.

\begin{table}[t]
\centering
\small
\begin{tabular}{@{}lr@{}}
\toprule
Statistic & Value \\
\midrule
Scenarios (SWE-bench / Lite) & 250 (200 / 50) \\
Distinct issues & 246 \\
Owners per scenario & 4 (100) or 5 (150) \\
Private files per scenario & 18.0 (16--20) \\
Private text per scenario (chars) & 15.5K (9.6K--21.9K) \\
SQLite tables / rows per owner & 2.5 / 3.6 \\
Owner constraints per scenario & 9.2 (8--10) \\
Requirement links per scenario & 29.4 (26--32) \\
Canonical actions per scenario & 11.6 (11--12) \\
Workflow predicates per scenario & 10 \\
Fixture files / lines & median 4 / 116 \\
\bottomrule
\end{tabular}
\caption{Corpus statistics. Ranges in parentheses.}
\label{tab:stats}
\end{table}

\paragraph{Corpus statistics.}
The 250 scenarios come from 246 distinct issues, because four Lite issues also seed a scenario in another archetype. They span ten repositories, led by Django (164) and Astropy (57), and involve 1,150 owners. Eight constraint kinds appear in every scenario, and the QA, security, and release archetypes add two kinds each. The reference patch touches one file in 153 scenarios and 2 to 22 files in the other 97. Table~\ref{tab:stats} summarizes the corpus, and Appendix~\ref{app:bundle} lists the artifacts of a bundle.

\section{Evaluation Protocol}
\label{sec:evaluation}

\bench reports three outcomes for every run and never folds one into another.

\subsection{Task Success}
\label{sec:ts}

Task success (\TS) asks whether the team's patch fixes the issue. We export the difference between the final and initial fixture as a SWE-bench prediction and run the official harness at the instance's base commit on the SWE-bench test split, which contains the Lite instances. A run succeeds when the harness marks the instance resolved: every FAIL\_TO\_PASS test passes and no PASS\_TO\_PASS test breaks. Empty and non-applying patches count as failures. \TS does not depend on whether the team submitted the patch or completed the workflow, and the harness receives each patch exactly as exported. Appendix~\ref{app:harness} describes the harness configuration and a gold-patch check.

\subsection{Workflow Success}

Workflow success (\WS) asks whether the team completed the process that the organization requires. Two conditions must hold. First, each canonical action of the contract, 11 or 12 per scenario, appears in the trace as a successful call by the responsible agent: the base owners' evidence reads, the implementer's write, the reviewer's approval, the code owner's grant, the gate owner's sign-off where the archetype has one, and the CI owner's test records and submission. Second, the workflow predicates hold on the final state: required files are changed and forbidden files untouched, the required approval covers the required files, target and regression runs are recorded as passing, the patch is submitted, and an evidence chain links the submission to the owner-evidence reads, the review decision, the approval, and the test records (the \emph{evidence} check). Test records attest to process: \texttt{repo.run\_tests} logs the CI owner's run without executing the upstream suite, and only the harness judges correctness.

\subsection{Constraint Success}

Constraint success (\CS) asks whether the team respected what individual owners require of the change. We check each owner's constraints against the final state and the trace. Patch-boundary and forbidden-file constraints (the \emph{boundary} check) hold when every required file is changed and no forbidden file is. Approval-scope and stale-approval constraints (the \emph{authority} check) hold when the responsible owner granted an approval in this run that covers the required files. The runtime accepts a grant only after the code owner's approval policy has been read and its ownership scope queried, and it attaches the owner's required files to every grant. Review and test obligations are enforced through \WS. Archetype-specific constraints that need semantic judgment, such as a performance-regression risk, are left to an optional LLM judge \citep{zheng2023judging} that never modifies \TS, \WS, or \CS (Appendix~\ref{app:harness}). Compliant success is the conjunction of \TS and \CS.

\paragraph{Constraint kinds.}
The eight base kinds are checked as follows. \emph{Patch boundary} and \emph{forbidden files} (implementer) are checked against the changed files; \emph{approval scope} and \emph{stale approval} (code owner) are checked against the approvals granted in the run; \emph{fresh review} and \emph{broad refactor rejected} (reviewer) and \emph{target tests} and \emph{regression tests} (CI owner) are enforced by \WS. The archetype-specific kinds are \emph{reproduction required} and \emph{acceptance scope} (QA), \emph{security exception} and \emph{secret files forbidden} (security), and \emph{release gate} and \emph{no performance regression} (SRE). They are linked to judge rubric axes and are not part of \CS in this paper.

\paragraph{Valid runs.}
Model-attributable failures count as failures: unparsable or empty replies, misaddressed messages, calls to tools outside the agent's role, and exhausted turn budgets. Runs that end for reasons unrelated to the model are retried, runs still lacking a valid outcome leave the denominator, and a completed run is never rerun.

\FloatBarrier
\section{Experiments}
\label{sec:experiments}

\begin{table}[t]
\centering
\small
\begin{tabular}{@{}lrrrrrrrr@{}}
\toprule
 & & \multicolumn{3}{c}{Outcomes} & \multicolumn{4}{c}{Diagnostic checks} \\
\cmidrule(lr){3-5}\cmidrule(l){6-9}
Model & Runs & \TS & \WS & \CS & Abort $\downarrow$ & Evidence & Authority & Boundary \\
\midrule
\multicolumn{9}{@{}l}{\emph{Frontier sweep}} \\
GPT-6.1 Sol & 250 & 35.2 & 78.4 & 98.8 & 2.0 & 98.8 & 99.2 & 99.6 \\
Claude Opus 5.5 & 250 & 20.8 & 47.2 & 90.8 & 50.4 & 69.6 & 92.8 & 98.0 \\
DeepSeek-V3.2 & 250 & 13.6 & 53.2 & 86.4 & 19.6 & 84.4 & 96.0 & 88.8 \\
\midrule
\multicolumn{9}{@{}l}{\emph{Earlier sweep}} \\
Claude Opus 4.7 & 250 & 54.0$^{\ast}$ & 84.8 & 85.6 & 0.0 & 84.8 & 98.0 & 87.6 \\
Claude Sonnet 4.5 & 220 & -- & 74.1 & 74.5 & 3.2 & 74.5 & 87.7 & 83.2 \\
Qwen3-32B & 250 & -- & 4.4 & 37.6 & 21.6 & 6.0 & 53.2 & 60.0 \\
\bottomrule
\end{tabular}
\caption{Results of both sweeps, in percent of valid runs. \TS: official harness. Frontier \WS, \CS, and checks are computed from trajectories (Appendix~\ref{app:replay}). \emph{Abort}: model-attributable failure or exhausted turn budget (the only kind in the frontier sweep). \emph{Evidence}, \emph{Authority}, \emph{Boundary}: the checks of Section~\ref{sec:evaluation}. --: not scored. $\ast$: Lite scenarios only.}
\label{tab:main}
\end{table}

\begin{figure}[t]
\centering
\includegraphics[width=\textwidth]{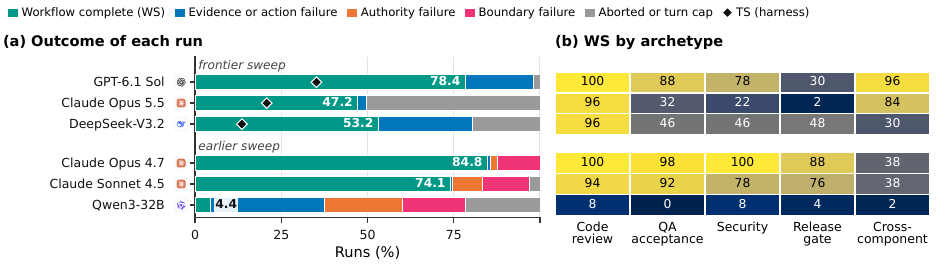}
\caption{Outcomes of every model in both sweeps. (a)~Each valid run is assigned one outcome: complete (\WS holds), aborted (including an exhausted turn budget), or its first failed check among boundary, authority, and evidence, where the last also covers missing canonical actions, such as a gate owner's sign-off, and missing test records. Labels give \WS, and diamonds give harness \TS. (b)~\WS by archetype (run counts in Tables~\ref{tab:archetype-all} and~\ref{tab:frontier-arch}). Frontier gate owners worked without their tables (\S\ref{sec:setup}).}
\label{fig:results}
\end{figure}

\begin{figure}[!t]
\centering
\includegraphics[width=0.6\linewidth]{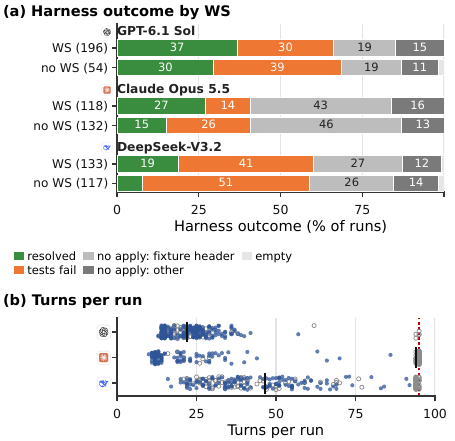}
\caption{Frontier sweep. (a)~Harness outcomes of runs with and without \WS (run counts in parentheses); non-applying patches are split by whether the diff includes the fixture header (\S\ref{sec:gap}). (b)~Turns per run: filled dots completed the workflow, bars mark medians, and the dashed line marks the turn cap.}
\label{fig:frontier}
\end{figure}

\subsection{Setup}
\label{sec:setup}

We evaluate homogeneous teams, in which every agent uses the same model, in two sweeps over the same scenarios and the Docker-based personal-workspace runtime of \citet{wang2026weclawarena}. Each scenario runs in isolated containers, the runtime routes and logs every message and tool call, and a run may take 90 turns plus one per agent. Each model runs once per scenario. The \emph{frontier sweep} covers GPT-6.1 Sol, Claude Opus 5.5, and DeepSeek-V3.2 \citep{deepseekai2025v32} on all 250 scenarios, with \TS from the official harness (version 3.0.17) for every run. Its trajectories record every message, tool call, and tool observation, and we compute \WS, \CS, and the diagnostic checks from them (Appendix~\ref{app:replay}). The \emph{earlier sweep} covers Claude Opus 4.7, Claude Sonnet 4.5, and Qwen3-32B \citep{yang2025qwen3}, served through Amazon Bedrock. It scores \WS, \CS, and the diagnostic checks, and \TS for Opus 4.7 on the 50 Lite scenarios (Appendix~\ref{app:accounting}). \todo{Rerun the credential-failed scenarios of the earlier sweep.}

\paragraph{Gate-owner evidence.}
The gated archetypes ran with less gate-owner evidence than their bundles specify. In the frontier sweep, the runtime did not serve the QA, security, and SRE owners' tables, so these owners worked from their notes alone (\S\ref{sec:where}). In both sweeps, the release-gate personas also refer the SRE owner to a rollout-policy file that the bundles do not contain. \todo{Rerun the gated archetypes with readable gate tables and a fixed SRE persona. Regenerate the released QA acceptance rows, which score every QA run as incomplete; the paper reports \WS computed from trajectories.}

\subsection{Main Results}

\begin{figure}[t]
\centering
\includegraphics[width=0.6\linewidth]{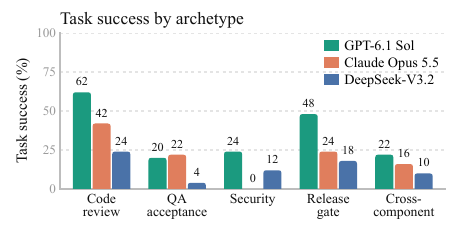}
\caption{Task success of the frontier agents by archetype, 50 scenarios per bar. Table~\ref{tab:frontier-arch} also gives \WS and \CS by archetype.}
\label{fig:ts-arch}
\end{figure}

\paragraph{The workflow separates models.}
In the earlier sweep, Opus 4.7 completes the workflow in 84.8\% of scenarios without breaking the interaction protocol, Sonnet 4.5 in 74.1\%, and Qwen3-32B in 4.4\% (Table~\ref{tab:main} and Figure~\ref{fig:results}). In the frontier sweep, GPT-6.1 Sol leads on both axes (35.2\% \TS and 78.4\% \WS). Opus 5.5 and DeepSeek-V3.2 complete the workflow at similar rates (47.2\% and 53.2\%, inside each other's 95\% intervals), but Opus 5.5 resolves more issues (20.8\% against 13.6\%). Code review is the easiest archetype, with at least 96\% \WS for every frontier agent and \TS of 62\%, 42\%, and 24\% (Figure~\ref{fig:ts-arch}). Only 18 scenarios are resolved by all three frontier teams, and 104 by at least one.

\begin{table}[!t]
\centering
\footnotesize
\setlength{\tabcolsep}{4.5pt}
\begin{tabular}{@{}lccccc@{}}
\toprule
Agent & Code review & QA acceptance & Security & Release gate & Cross-component \\
\midrule
GPT-6.1 Sol & 62 / 100 / 100 & 20 / 88 / 100 & 24 / 78 / 100 & 48 / 30 / 98 & 22 / 96 / 96 \\
Claude Opus 5.5 & 42 / 96 / 100 & 22 / 32 / 84 & 0 / 22 / 84 & 24 / 2 / 96 & 16 / 84 / 90 \\
DeepSeek-V3.2 & 24 / 96 / 100 & 4 / 46 / 100 & 12 / 46 / 96 & 18 / 48 / 98 & 10 / 30 / 38 \\
\bottomrule
\end{tabular}
\caption{Task, workflow, and constraint success (\TS / \WS / \CS, \%) of the frontier sweep by archetype, 50 runs per cell. \WS and \CS are computed from trajectories (Appendix~\ref{app:replay}); frontier gate owners worked without their tables (\S\ref{sec:setup}).}
\label{tab:frontier-arch}
\end{table}

\paragraph{Constraints fail by omission.}
No run in either sweep edited the decoy secrets file, which the personas name explicitly, and most teams obtain the code owner's approval (92.8--99.2\% of frontier runs). The binding constraint is the patch boundary: all 31 boundary failures of Opus 4.7 and all 34 frontier boundary failures leave required files unchanged. Qwen3-32B shows a different profile: its \CS (37.6\%) far exceeds its \WS (4.4\%), because its teams often stay inside the boundary and obtain approval but never complete the process.

\subsection{Workflow Completion Is Not Correctness}
\label{sec:gap}

\WS requires a patch that has been reviewed, approved within scope, and signed off by the CI owner, so one might expect it to imply a correct patch. Yet among workflow-complete runs, the harness resolves only 36.7\% for GPT-6.1 Sol, 27.1\% for Opus 5.5, and 18.8\% for DeepSeek-V3.2; the others ship a patch that fails the upstream tests or does not apply (Figure~\ref{fig:frontier}a). Conversely, 16, 20, and 9 resolved patches (18\%, 38\%, and 26\%) come from runs that never completed the workflow, each because a gate owner never signed off or the team ran out of turns. Crediting the workflow as task success would overstate resolution by a factor of 2.2 to 3.9. The earlier sweep shows the same gap: all 50 Opus 4.7 runs on the Lite scenarios completed the workflow and satisfied every checked constraint, yet the harness resolves 27 of their patches (Table~\ref{tab:lite-repo}).

Many unresolved patches never reach the tests. Each reconstructed source file in a fixture opens with a two-line comment, the \emph{fixture header}, which does not exist upstream. The diff includes lines of it in 225 of the 328 non-applying frontier patches (47 of 82, 112 of 148, and 66 of 98) and in 10 of the 14 non-applying Opus 4.7 Lite patches, and such a patch does not apply. \todo{Rescore the header-carrying patches with the header stripped.}

\subsection{Where Teams Fail}
\label{sec:where}

\paragraph{Missing evidence stalls gated reviews.}
In the gated archetypes, the QA, security, or SRE owner must sign off, and in the frontier sweep it had only its notes to go on (\S\ref{sec:setup}). The Opus 5.5 gate owner withholds sign-off in 120 of 150 gated runs, and the team waits until the turn budget runs out in 114 of them, although its completed workflows need a median of only 13 turns (Figure~\ref{fig:frontier}b); 20 of its capped runs hold a resolved patch. The GPT-6.1 Sol and DeepSeek-V3.2 gate owners withhold sign-off less often (52 and 68 runs), and in 49 and 64 of those runs the CI owner submits anyway. Both fail \WS, one by stalling and the other by shipping without a required sign-off. Runs that fail \WS this way can still resolve the issue: on release-gate scenarios, GPT-6.1 Sol and Opus 5.5 resolve 48\% and 24\% of the issues while completing 30\% and 2\% of the workflows (Figure~\ref{fig:ts-arch}). In the earlier sweep, Opus 4.7 completed 98\%, 100\%, and 88\% of these workflows.

\paragraph{Cross-component ownership is the hardest.}
In the earlier sweep, Opus 4.7 completes 88--100\% of the single-scope archetypes but 38\% of cross-component changes, and Sonnet 4.5 drops from 76--94\% to 38\% (Figure~\ref{fig:results}b). These are boundary failures that grow with the number of required files: Opus 4.7 completes 45\%, 33\%, and 20\% of the scenarios with two, three, and four or more. Boundary failures track what the owners' records name. In 40 of the 50 cross-component bundles, the notes and tables name only some of the required files, although the opening message lists all of them. In 25 of the 33 frontier boundary failures in this archetype, every unchanged file is one that the records omit. DeepSeek-V3.2 completes 30\% of these workflows and exhausts the turn budget in the other 35, in 27 of them while its CI owner retries a submission refused for unchanged required files. GPT-6.1 Sol and Opus 5.5 complete 96\% and 84\% of these workflows but resolve only 22\% and 16\% (Figure~\ref{fig:ts-arch}; 38.5\% and 22.0\% on single-scope scenarios).

\paragraph{Weak and strong teams fail differently.}
The dominant outcome in Figure~\ref{fig:results}a shifts with capability. Qwen3-32B aborts one run in five through malformed, misaddressed, or disallowed actions, and it submits in 181 scored runs without the reads and records that the contract requires. The strongest teams fail at the edges of their owners' evidence, at files the records omit and at sign-offs that gate owners could not ground. Each level calls for a different remedy: reliable action formatting, evidence discipline, or reasoning about missing evidence.

\FloatBarrier
\section{Conclusion}

\bench tests whether coding agents acting for different people can ship a correct patch through the process those people require. Across six models, executing the process separates teams, a completed workflow says little about correctness, and teams falter where an owner's evidence is incomplete. Our code and benchmark are available at \url{https://github.com/kingofspace0wzz/SWE-Team-Bench}.

\section{Limitations}
\label{sec:limitations}

Agents edit reduced fixtures rather than full checkouts, which makes the coding subproblem easier than in SWE-bench, although the official harness still judges every patch against the full repository. Personas state the workflow explicitly, so \bench measures faithful execution of a known process under partitioned authority rather than the discovery of unwritten norms. One LLM writes all owner resources under fixed blueprints, so they can be formulaic, and the boundary check expects the files of the reference patch. We evaluate homogeneous teams with one run per scenario, and Django accounts for 164 of the 250 scenarios.

\section{Ethics Statement}
\label{sec:ethics}

\bench is built from public SWE-bench instances. Personas, notes, records, and the decoy secrets file are synthetic and contain no personal data; the decoy credential is a placeholder string. The benchmark measures whether agents respect ownership boundaries, which can help developers of agent teams find scope and approval failures before deployment. All models were accessed through a commercial API under its terms of use.

\bibliography{main}

\begin{thebibliography}{45}
\providecommand{\natexlab}[1]{#1}
\providecommand{\url}[1]{\texttt{#1}}
\expandafter\ifx\csname urlstyle\endcsname\relax
  \providecommand{\doi}[1]{doi: #1}\else
  \providecommand{\doi}{doi: \begingroup \urlstyle{rm}\Url}\fi

\bibitem[Abdelnabi et~al.(2025)Abdelnabi, Gomaa, Bagdasarian, Kristensson, and
  Shokri]{abdelnabi2025firewalls}
Sahar Abdelnabi, Amr Gomaa, Eugene Bagdasarian, Per~Ola Kristensson, and Reza
  Shokri.
\newblock Firewalls to secure dynamic {LLM} agentic networks.
\newblock \emph{arXiv preprint arXiv:2502.01822}, 2025.

\bibitem[Bacchelli \& Bird(2013)Bacchelli and Bird]{bacchelli2013expectations}
Alberto Bacchelli and Christian Bird.
\newblock Expectations, outcomes, and challenges of modern code review.
\newblock In \emph{Proceedings of the 35th International Conference on Software
  Engineering (ICSE)}, pp.\  712--721. IEEE, 2013.
\newblock \doi{10.1109/ICSE.2013.6606617}.

\bibitem[Barres et~al.(2025)Barres, Dong, Ray, Si, and
  Narasimhan]{barres2025tau2bench}
Victor Barres, Honghua Dong, Soham Ray, Xujie Si, and Karthik Narasimhan.
\newblock {$\tau^2$-Bench}: Evaluating conversational agents in a dual-control
  environment.
\newblock \emph{arXiv preprint arXiv:2506.07982}, 2025.

\bibitem[Bird et~al.(2011)Bird, Nagappan, Murphy, Gall, and
  Devanbu]{bird2011dont}
Christian Bird, Nachiappan Nagappan, Brendan Murphy, Harald Gall, and Premkumar
  Devanbu.
\newblock Don't touch my code! examining the effects of ownership on software
  quality.
\newblock In \emph{Proceedings of the 19th ACM SIGSOFT Symposium and the 13th
  European Conference on Foundations of Software Engineering}, pp.\  4--14.
  ACM, 2011.
\newblock \doi{10.1145/2025113.2025119}.

\bibitem[Cai et~al.(2026)Cai, Lyu, Ni, Chen, Zhou, Zhu, Lu, Wang, Ruan,
  Schneider, Zhang, Li, Zheng, Zhang, Nie, and Chen]{cai2026sweqapro}
Songcheng Cai, Zhiheng Lyu, Yuansheng Ni, Xiangchao Chen, Baichuan Zhou,
  Shenzhe Zhu, Yi~Lu, Haozhe Wang, Chi Ruan, Benjamin Schneider, Weixu Zhang,
  Xiang Li, Andy Zheng, Yuyu Zhang, Ping Nie, and Wenhu Chen.
\newblock {SWE-QA-Pro}: A representative benchmark and scalable training recipe
  for repository-level code understanding.
\newblock \emph{arXiv preprint arXiv:2603.16124}, 2026.

\bibitem[Cemri et~al.(2025)Cemri, Pan, Yang, Agrawal, Chopra, Tiwari, Keutzer,
  Parameswaran, Klein, Ramchandran, Zaharia, Gonzalez, and
  Stoica]{cemri2025multiagent}
Mert Cemri, Melissa~Z. Pan, Shuyi Yang, Lakshya~A. Agrawal, Bhavya Chopra,
  Rishabh Tiwari, Kurt Keutzer, Aditya Parameswaran, Dan Klein, Kannan
  Ramchandran, Matei Zaharia, Joseph~E. Gonzalez, and Ion Stoica.
\newblock Why do multi-agent {LLM} systems fail?
\newblock In \emph{Advances in Neural Information Processing Systems}, 2025.

\bibitem[Chen et~al.(2024)Chen, Su, Zuo, Yang, Yuan, Chan, Yu, Lu, Hung, Qian,
  et~al.]{chen2024agentverse}
Weize Chen, Yusheng Su, Jingwei Zuo, Cheng Yang, Chenfei Yuan, Chi-Min Chan,
  Heyang Yu, Yaxi Lu, Yi-Hsin Hung, Chen Qian, et~al.
\newblock Agent{V}erse: Facilitating multi-agent collaboration and exploring
  emergent behaviors.
\newblock In \emph{The Twelfth International Conference on Learning
  Representations}, 2024.

\bibitem[{DeepSeek-AI}(2025)]{deepseekai2025v32}
{DeepSeek-AI}.
\newblock {DeepSeek-V3.2}: Pushing the frontier of open large language models.
\newblock \emph{arXiv preprint arXiv:2512.02556}, 2025.

\bibitem[Deng et~al.(2026)Deng, Chen, Yu, Fan, Liu, Yang, Parikh, Kannan, Cong,
  Wang, Zhang, Prasanna, Tang, and Wang]{deng2026swemilestone}
Gangda Deng, Zhaoling Chen, Zhongming Yu, Haoyang Fan, Yuhong Liu, Yuxin Yang,
  Dhruv Parikh, Rajgopal Kannan, Le~Cong, Mengdi Wang, Qian Zhang, Viktor
  Prasanna, Xiangru Tang, and Xingyao Wang.
\newblock {SWE-Milestone}: Evaluating {AI} agents on continuous software
  evolution.
\newblock \emph{arXiv preprint arXiv:2603.13428}, 2026.
\newblock Earlier versions were titled \emph{EvoClaw}.

\bibitem[Deng et~al.(2025)Deng, Da, Pan, He, Ide, Garg, Lauffer, Park, Pasari,
  Rane, Sampath, Krishnan, Kundurthy, Hendryx, Wang, Bharadwaj, Holm, Aluri,
  Zhang, Jacobson, Liu, and Kenstler]{deng2025swebenchpro}
Xiang Deng, Jeff Da, Edwin Pan, Yannis~Yiming He, Charles Ide, Kanak Garg,
  Niklas Lauffer, Andrew Park, Nitin Pasari, Chetan Rane, Karmini Sampath, Maya
  Krishnan, Srivatsa Kundurthy, Sean Hendryx, Zifan Wang, Vijay Bharadwaj, Jeff
  Holm, Raja Aluri, Chen Bo~Calvin Zhang, Noah Jacobson, Bing Liu, and Brad
  Kenstler.
\newblock {SWE-Bench Pro}: Can {AI} agents solve long-horizon software
  engineering tasks?
\newblock \emph{arXiv preprint arXiv:2509.16941}, 2025.

\bibitem[Drouin et~al.(2024)Drouin, Gasse, Caccia, Laradji, Del~Verme, Marty,
  Boisvert, Thakkar, Cappart, Vazquez, Chapados, and
  Lacoste]{drouin2024workarena}
Alexandre Drouin, Maxime Gasse, Massimo Caccia, Issam~H. Laradji, Manuel
  Del~Verme, Tom Marty, L{\'e}o Boisvert, Megh Thakkar, Quentin Cappart, David
  Vazquez, Nicolas Chapados, and Alexandre Lacoste.
\newblock {WorkArena}: How capable are web agents at solving common knowledge
  work tasks?
\newblock In \emph{International Conference on Machine Learning}, 2024.

\bibitem[He et~al.(2025)He, Treude, and Lo]{he2025llmmas}
Junda He, Christoph Treude, and David Lo.
\newblock {LLM}-based multi-agent systems for software engineering: Literature
  review, vision, and the road ahead.
\newblock \emph{ACM Transactions on Software Engineering and Methodology},
  34\penalty0 (5), 2025.
\newblock \doi{10.1145/3712003}.

\bibitem[Hong et~al.(2024)Hong, Zhuge, Chen, Zheng, Cheng, Wang, Zhang, Wang,
  Lin, Zhou, Ran, Xiao, Wu, and Schmidhuber]{hong2024metagpt}
Sirui Hong, Mingchen Zhuge, Jonathan Chen, Xiawu Zheng, Yuheng Cheng, Ceyao
  Wang, Zili Zhang, Steven Ka Shing~Yau Wang, Zijuan Lin, Liyang Zhou, Chenyu
  Ran, Lingfeng Xiao, Chenglin Wu, and J{\"u}rgen Schmidhuber.
\newblock {MetaGPT}: Meta programming for a multi-agent collaborative
  framework.
\newblock In \emph{The Twelfth International Conference on Learning
  Representations}, 2024.

\bibitem[Jimenez et~al.(2024)Jimenez, Yang, Wettig, Yao, Pei, Press, and
  Narasimhan]{jimenez2024swebench}
Carlos~E. Jimenez, John Yang, Alexander Wettig, Shunyu Yao, Kexin Pei, Ofir
  Press, and Karthik~R. Narasimhan.
\newblock {SWE}-bench: Can language models resolve real-world {GitHub} issues?
\newblock In \emph{The Twelfth International Conference on Learning
  Representations}, 2024.
\newblock URL \url{https://openreview.net/forum?id=VTF8yNQM66}.

\bibitem[Juneja et~al.(2025)Juneja, Pasupulati, Albalak, Hua, and
  Wang]{juneja2025magpie}
Gurusha Juneja, Jayanth Naga~Sai Pasupulati, Alon Albalak, Wenyue Hua, and
  William~Yang Wang.
\newblock {MAGPIE}: A benchmark for multi-{AG}ent contextual {PrI}vacy
  {E}valuation.
\newblock \emph{arXiv preprint arXiv:2510.15186}, 2025.

\bibitem[Levy et~al.(2024)Levy, Wiesel, Marreed, Oved, Yaeli, Mashkif, and
  Shlomov]{levy2024stwebagentbench}
Ido Levy, Ben Wiesel, Sami Marreed, Alon Oved, Avi Yaeli, Nir Mashkif, and
  Segev Shlomov.
\newblock {ST-WebAgentBench}: A benchmark for evaluating safety and
  trustworthiness in web agents.
\newblock \emph{arXiv preprint arXiv:2410.06703}, 2024.

\bibitem[Liang et~al.(2026)Liang, Lyu, Liu, Chen, Nie, Zou, and
  Chen]{liang2026swenext}
Jiarong Liang, Zhiheng Lyu, Zijie Liu, Xiangchao Chen, Ping Nie, Kai Zou, and
  Wenhu Chen.
\newblock {SWE-Next}: Scalable real-world software engineering tasks for
  agents.
\newblock \emph{arXiv preprint arXiv:2603.20691}, 2026.

\bibitem[Mireshghallah et~al.(2024)Mireshghallah, Kim, Zhou, Tsvetkov, Sap,
  Shokri, and Choi]{mireshghallah2024confaide}
Niloofar Mireshghallah, Hyunwoo Kim, Xuhui Zhou, Yulia Tsvetkov, Maarten Sap,
  Reza Shokri, and Yejin Choi.
\newblock Can {LLM}s keep a secret? testing privacy implications of language
  models via contextual integrity theory.
\newblock In \emph{International Conference on Learning Representations}, 2024.

\bibitem[Miserendino et~al.(2025)Miserendino, Wang, Patwardhan, and
  Heidecke]{miserendino2025swelancer}
Samuel Miserendino, Michele Wang, Tejal Patwardhan, and Johannes Heidecke.
\newblock {SWE-Lancer}: Can frontier {LLMs} earn \$1 million from real-world
  freelance software engineering?
\newblock In \emph{International Conference on Machine Learning}, 2025.

\bibitem[Nissenbaum(2004)]{nissenbaum2004privacy}
Helen Nissenbaum.
\newblock Privacy as contextual integrity.
\newblock \emph{Washington Law Review}, 79\penalty0 (1):\penalty0 119--158,
  2004.

\bibitem[Pan et~al.(2025)Pan, Wang, Neubig, Jaitly, Ji, Suhr, and
  Zhang]{pan2025swegym}
Jiayi Pan, Xingyao Wang, Graham Neubig, Navdeep Jaitly, Heng Ji, Alane Suhr,
  and Yizhe Zhang.
\newblock Training software engineering agents and verifiers with {SWE-Gym}.
\newblock In \emph{International Conference on Machine Learning}, 2025.

\bibitem[Qian et~al.(2024)Qian, Liu, Liu, Chen, Dang, Li, Yang, Chen, Su, Cong,
  Xu, Li, Liu, and Sun]{qian2024chatdev}
Chen Qian, Wei Liu, Hongzhang Liu, Nuo Chen, Yufan Dang, Jiahao Li, Cheng Yang,
  Weize Chen, Yusheng Su, Xin Cong, Juyuan Xu, Dahai Li, Zhiyuan Liu, and
  Maosong Sun.
\newblock {C}hat{D}ev: Communicative agents for software development.
\newblock In \emph{Proceedings of the 62nd Annual Meeting of the Association
  for Computational Linguistics (Volume 1: Long Papers)}, pp.\  15174--15186,
  Bangkok, Thailand, 2024. Association for Computational Linguistics.
\newblock \doi{10.18653/v1/2024.acl-long.810}.

\bibitem[Qian et~al.(2025)Qian, Xie, Wang, Liu, Zhu, Xia, Dang, Du, Chen, Yang,
  Liu, and Sun]{qian2025scaling}
Chen Qian, Zihao Xie, YiFei Wang, Wei Liu, Kunlun Zhu, Hanchen Xia, Yufan Dang,
  Zhuoyun Du, Weize Chen, Cheng Yang, Zhiyuan Liu, and Maosong Sun.
\newblock Scaling large language model-based multi-agent collaboration.
\newblock In \emph{The Thirteenth International Conference on Learning
  Representations}, 2025.

\bibitem[Sadowski et~al.(2018)Sadowski, S{\"o}derberg, Church, Sipko, and
  Bacchelli]{sadowski2018modern}
Caitlin Sadowski, Emma S{\"o}derberg, Luke Church, Michal Sipko, and Alberto
  Bacchelli.
\newblock Modern code review: A case study at {Google}.
\newblock In \emph{Proceedings of the 40th International Conference on Software
  Engineering: Software Engineering in Practice}, pp.\  181--190. ACM, 2018.
\newblock \doi{10.1145/3183519.3183525}.

\bibitem[Shao et~al.(2024{\natexlab{a}})Shao, Li, Shi, Liu, and
  Yang]{shao2024privacylens}
Yijia Shao, Tianshi Li, Weiyan Shi, Yanchen Liu, and Diyi Yang.
\newblock {PrivacyLens}: Evaluating privacy norm awareness of language models
  in action.
\newblock In \emph{Advances in Neural Information Processing Systems, Datasets
  and Benchmarks Track}, 2024{\natexlab{a}}.

\bibitem[Shao et~al.(2024{\natexlab{b}})Shao, Samuel, Jiang, Yang, and
  Yang]{shao2024collaborativegym}
Yijia Shao, Vinay Samuel, Yucheng Jiang, John Yang, and Diyi Yang.
\newblock Collaborative gym: A framework for enabling and evaluating
  human-agent collaboration.
\newblock \emph{arXiv preprint arXiv:2412.15701}, 2024{\natexlab{b}}.

\bibitem[Steinberger \& {OpenClaw Contributors}(2025)Steinberger and {OpenClaw
  Contributors}]{openclaw2025}
Peter Steinberger and {OpenClaw Contributors}.
\newblock {OpenClaw}: Open-source agent framework.
\newblock \url{https://github.com/openclaw/openclaw}, 2025.

\bibitem[Tang et~al.(2024)Tang, Kim, Song, Lothritz, Li, Ezzini, Tian, Klein,
  and Bissyand{\'e}]{tang2024codeagent}
Xunzhu Tang, Kisub Kim, Yewei Song, Cedric Lothritz, Bei Li, Saad Ezzini, Haoye
  Tian, Jacques Klein, and Tegawend{\'e}~F. Bissyand{\'e}.
\newblock {C}ode{A}gent: Autonomous communicative agents for code review.
\newblock In \emph{Proceedings of the 2024 Conference on Empirical Methods in
  Natural Language Processing}, pp.\  11279--11313, Miami, Florida, USA, 2024.
  Association for Computational Linguistics.
\newblock \doi{10.18653/v1/2024.emnlp-main.632}.

\bibitem[Tang et~al.(2026)Tang, Zhou, Liu, Li, Wu, Wang, Huang, Zhou, Zhou,
  Song, Yu, Wang, Zhou, Zhou, Lv, Li, Liu, Chen, Liu, Li, Kang, and
  Wu]{tang2026workspacebench}
Zirui Tang, Xuanhe Zhou, Yumou Liu, Linchun Li, Yukai Wu, Weizheng Wang,
  Hongzhang Huang, Wei Zhou, Jun Zhou, Jiachen Song, Shaoli Yu, Jinqi Wang,
  Zihang Zhou, Hongyi Zhou, Yuting Lv, Jinyang Li, Jiashuo Liu, Ruoyu Chen,
  Chunwei Liu, GuoLiang Li, Jihua Kang, and Fan Wu.
\newblock {Workspace-Bench} 1.0: Benchmarking {AI} agents on workspace tasks
  with large-scale file dependencies.
\newblock \emph{arXiv preprint arXiv:2605.03596}, 2026.

\bibitem[Trivedi et~al.(2024)Trivedi, Khot, Hartmann, Manku, Dong, Li, Gupta,
  Sabharwal, and Balasubramanian]{trivedi2024appworld}
Harsh Trivedi, Tushar Khot, Mareike Hartmann, Ruskin Manku, Vinty Dong, Edward
  Li, Shashank Gupta, Ashish Sabharwal, and Niranjan Balasubramanian.
\newblock {AppWorld}: A controllable world of apps and people for benchmarking
  interactive coding agents.
\newblock In \emph{Proceedings of the 62nd Annual Meeting of the Association
  for Computational Linguistics}, 2024.

\bibitem[Wang \& Jiang(2026{\natexlab{a}})Wang and
  Jiang]{wang2026agentsocialbench}
Prince~Zizhuang Wang and Shuli Jiang.
\newblock {AgentSocialBench}: Evaluating privacy risks in human-centered
  agentic social networks.
\newblock In \emph{Proceedings of the 2026 Conference on Empirical Methods in
  Natural Language Processing}, pp.\  40763--40785, Budapest, Hungary,
  2026{\natexlab{a}}. Association for Computational Linguistics.
\newblock \doi{10.18653/2026.emnlp-main.1789}.

\bibitem[Wang \& Jiang(2026{\natexlab{b}})Wang and Jiang]{wang2026prime}
Prince~Zizhuang Wang and Shuli Jiang.
\newblock {PRIME}: Training free proactive reasoning via iterative memory
  evolution for user-centric agent.
\newblock \emph{arXiv preprint arXiv:2604.07645}, 2026{\natexlab{b}}.

\bibitem[Wang et~al.(2026)Wang, Yuan, Zhang, Hu, Zhao, and
  Jiang]{wang2026weclawarena}
Prince~Zizhuang Wang, Aojie Yuan, Haiyue Zhang, Xiyang Hu, Yue Zhao, and Shuli
  Jiang.
\newblock {WeClawArena}: An auditable sandbox and benchmark for cross-user
  agents collaboration and security in human-centered agent networks.
\newblock In \emph{Findings of the Association for Computational Linguistics:
  EMNLP 2026}, pp.\  30350--30377, Budapest, Hungary, 2026. Association for
  Computational Linguistics.
\newblock \doi{10.18653/2026.findings-emnlp.1424}.

\bibitem[Wang et~al.(2025)Wang, Li, Song, Xu, Tang, Zhuge, Pan, Song, Li,
  Singh, Tran, Li, Ma, Zheng, Qian, Shao, Muennighoff, Zhang, Hui, Lin,
  Brennan, Peng, Ji, and Neubig]{wang2025openhands}
Xingyao Wang, Boxuan Li, Yufan Song, Frank~F. Xu, Xiangru Tang, Mingchen Zhuge,
  Jiayi Pan, Yueqi Song, Bowen Li, Jaskirat Singh, Hoang~H. Tran, Fuqiang Li,
  Ren Ma, Mingzhang Zheng, Bill Qian, Yanjun Shao, Niklas Muennighoff, Yizhe
  Zhang, Binyuan Hui, Junyang Lin, Robert Brennan, Hao Peng, Heng Ji, and
  Graham Neubig.
\newblock {OpenHands}: An open platform for {AI} software developers as
  generalist agents.
\newblock In \emph{The Thirteenth International Conference on Learning
  Representations}, 2025.

\bibitem[Wu et~al.(2023)Wu, Bansal, Zhang, Wu, Li, Zhu, Jiang, Zhang, Zhang,
  Liu, et~al.]{wu2023autogen}
Qingyun Wu, Gagan Bansal, Jieyu Zhang, Yiran Wu, Beibin Li, Erkang Zhu,
  Li~Jiang, Xiaoyun Zhang, Shaokun Zhang, Jiale Liu, et~al.
\newblock Auto{G}en: Enabling next-gen {LLM} applications via multi-agent
  conversation.
\newblock \emph{arXiv preprint arXiv:2308.08155}, 2023.

\bibitem[Xia et~al.(2024)Xia, Deng, Dunn, and Zhang]{xia2024agentless}
Chunqiu~Steven Xia, Yinlin Deng, Soren Dunn, and Lingming Zhang.
\newblock Agentless: Demystifying {LLM}-based software engineering agents.
\newblock \emph{arXiv preprint arXiv:2407.01489}, 2024.

\bibitem[Xu et~al.(2025)Xu, Song, Li, Tang, Jain, Bao, Wang, Zhou, Guo, Cao,
  Yang, Lu, Martin, Su, Maben, Mehta, Chi, Jang, Xie, Zhou, and
  Neubig]{xu2025theagentcompany}
Frank~F. Xu, Yufan Song, Boxuan Li, Yuxuan Tang, Kritanjali Jain, Mengxue Bao,
  Zora~Z. Wang, Xuhui Zhou, Zhitong Guo, Murong Cao, Mingyang Yang, Hao~Yang
  Lu, Amaad Martin, Zhe Su, Leander Maben, Raj Mehta, Wayne Chi, Lawrence Jang,
  Yiqing Xie, Shuyan Zhou, and Graham Neubig.
\newblock {TheAgentCompany}: Benchmarking {LLM} agents on consequential real
  world tasks.
\newblock In \emph{Advances in Neural Information Processing Systems}, 2025.

\bibitem[Yang et~al.(2025{\natexlab{a}})Yang, Li, Yang, Zhang, Hui, Zheng, Yu,
  Gao, Huang, Lv, et~al.]{yang2025qwen3}
An~Yang, Anfeng Li, Baosong Yang, Beichen Zhang, Binyuan Hui, Bo~Zheng, Bowen
  Yu, Chang Gao, Chengen Huang, Chenxu Lv, et~al.
\newblock Qwen3 technical report.
\newblock \emph{arXiv preprint arXiv:2505.09388}, 2025{\natexlab{a}}.

\bibitem[Yang et~al.(2024)Yang, Jimenez, Wettig, Lieret, Yao, Narasimhan, and
  Press]{yang2024sweagent}
John Yang, Carlos~E. Jimenez, Alexander Wettig, Kilian Lieret, Shunyu Yao,
  Karthik Narasimhan, and Ofir Press.
\newblock {SWE}-agent: Agent-computer interfaces enable automated software
  engineering.
\newblock In \emph{Advances in Neural Information Processing Systems}, 2024.

\bibitem[Yang et~al.(2025{\natexlab{b}})Yang, Lieret, Jimenez, Wettig,
  Khandpur, Zhang, Hui, Press, Schmidt, and Yang]{yang2025swesmith}
John Yang, Kilian Lieret, Carlos~E. Jimenez, Alexander Wettig, Kabir Khandpur,
  Yanzhe Zhang, Binyuan Hui, Ofir Press, Ludwig Schmidt, and Diyi Yang.
\newblock {SWE-smith}: Scaling data for software engineering agents.
\newblock In \emph{Advances in Neural Information Processing Systems},
  2025{\natexlab{b}}.

\bibitem[Yang et~al.(2025{\natexlab{c}})Yang, Lieret, Yang, Jimenez, Oblokulov,
  Siddiqui, Press, Schmidt, and Yang]{yang2025codeclash}
John Yang, Kilian Lieret, Joyce Yang, Carlos~E. Jimenez, Muhtasham Oblokulov,
  Aryan Siddiqui, Ofir Press, Ludwig Schmidt, and Diyi Yang.
\newblock {CodeClash}: Benchmarking goal-oriented software engineering.
\newblock \emph{arXiv preprint arXiv:2511.00839}, 2025{\natexlab{c}}.

\bibitem[Yao et~al.(2025)Yao, Shinn, Razavi, and Narasimhan]{yao2024taubench}
Shunyu Yao, Noah Shinn, Pedram Razavi, and Karthik Narasimhan.
\newblock {$\tau$-bench}: A benchmark for tool-agent-user interaction in
  real-world domains.
\newblock In \emph{International Conference on Learning Representations}, 2025.
\newblock arXiv:2406.12045.

\bibitem[Zan et~al.(2025)Zan, Huang, Liu, Chen, Zhang, Xin, Chen, Liu, Zhong,
  Li, Liu, Xiao, Chen, Zhang, Su, Liu, Long, Shen, and
  Xiang]{zan2025multiswebench}
Daoguang Zan, Zhirong Huang, Wei Liu, Hanwu Chen, Linhao Zhang, Shulin Xin,
  Lu~Chen, Qi~Liu, Xiaojian Zhong, Aoyan Li, Siyao Liu, Yongsheng Xiao,
  Liangqiang Chen, Yuyu Zhang, Jing Su, Tianyu Liu, Rui Long, Kai Shen, and
  Liang Xiang.
\newblock {Multi-SWE-bench}: A multilingual benchmark for issue resolving.
\newblock In \emph{Advances in Neural Information Processing Systems}, 2025.

\bibitem[Zheng et~al.(2023)Zheng, Chiang, Sheng, Zhuang, Wu, Zhuang, Lin, Li,
  Li, Xing, Zhang, Gonzalez, and Stoica]{zheng2023judging}
Lianmin Zheng, Wei-Lin Chiang, Ying Sheng, Siyuan Zhuang, Zhanghao Wu, Yonghao
  Zhuang, Zi~Lin, Zhuohan Li, Dacheng Li, Eric~P Xing, Hao Zhang, Joseph~E
  Gonzalez, and Ion Stoica.
\newblock Judging {LLM}-as-a-judge with {MT}-bench and chatbot arena.
\newblock In \emph{Advances in Neural Information Processing Systems},
  volume~36, 2023.

\bibitem[Zhu et~al.(2025)Zhu, Du, Hong, Yang, Guo, Wang, Wang, Qian, Tang, Ji,
  and You]{zhu2025multiagentbench}
Kunlun Zhu, Hongyi Du, Zhaochen Hong, Xiaocheng Yang, Shuyi Guo, Zhe Wang,
  Zhenhailong Wang, Cheng Qian, Xiangru Tang, Heng Ji, and Jiaxuan You.
\newblock {MultiAgentBench}: Evaluating the collaboration and competition of
  {LLM} agents.
\newblock \emph{arXiv preprint arXiv:2503.01935}, 2025.

\end{thebibliography}
\bibliographystyle{iclr2027_conference}

\clearpage
\appendix

\section{Scenario Bundle Artifacts}
\label{app:bundle}

\begin{table}[!ht]
\centering
\footnotesize
\begin{tabular}{@{}p{0.30\linewidth}p{0.64\linewidth}@{}}
\toprule
Artifact & Contents \\
\midrule
\texttt{scenario.json} & Agents, owners, persona prompts, communication graph, allowed tools, turn budget, opening message. \\
\texttt{scenario.gt.json} & Canonical actions, workflow predicates, governance and resource manifests, construction records. \\
\texttt{phase5\_metadata.json} & Source instance and dataset, archetype, required evidence, task labels. \\
\texttt{personas/} & Per-agent system prompts and tool allowlists. \\
\texttt{resources/owners/} & Per-owner notes, SQLite database, and \texttt{constraints.json}. \\
\texttt{resources/repo/} & Reduced fixture, target and regression test lists, repository metadata. \\
\texttt{resources/team/} & Ownership map, review policy, CI policy. \\
\texttt{mcp/} & Declarations of the 13 tools. \\
\bottomrule
\end{tabular}
\caption{Main artifacts of a scenario bundle.}
\label{tab:bundle}
\end{table}

\section{Construction Details}
\label{app:construction}

\paragraph{Task fact packs and blueprints.}
For each SWE-bench instance, we extract the issue, repository, base commit, FAIL\_TO\_PASS and PASS\_TO\_PASS tests, the files the reference patch touches, and bounded code excerpts. Reference-patch metadata determines the required files, the approval scope, and the validation targets, but it is never copied into owner resources as a recipe. A versioned blueprint for each archetype fixes the required roles, the private files and SQLite tables that each role must hold (with minimum row counts), the constraint objects, and the predicate or judge axis that checks each requirement.

\paragraph{Role-specific writing.}
For each owner, a deterministic step drafts a role-specific scaffold of files, tables, and constraints, and an LLM writer enriches it with task-specific evidence from the fact pack. The fact pack names the files that the reference patch touches but does not contain the patch, and the writer is instructed not to produce one. Each bundle also carries a requirement map that links every private resource to the predicates and judge axes that evaluate it, with 29.4 links per scenario on average.

\paragraph{Validation.}
Bundles pass four gates. Static validation checks schemas, persona and tool consistency, resource paths, and table row counts. Resource-quality validation rejects role-only templates, empty constraint files, and private notes that lack scenario-specific identifiers. Coverage validation requires every resource and constraint to be checked by at least one workflow predicate, constraint predicate, harness field, or judge axis. Runtime validation runs mock and live smoke tests and requires traces, scorecards, and an exportable patch. An LLM critic reviews a compact summary of each bundle and may trigger a bounded number of repairs, but it cannot override a deterministic failure. Each bundle records the fact-pack hash, blueprint version, writer and critic models, the critic's verdict, and the number of repairs; the critic accepted all 250 released bundles without repair. \todo{All 50 release-gate personas instruct the SRE owner to read \texttt{sre/rollout\_policy.md}, which does not exist in the SRE owner's files. Fix the bundles and rerun this archetype.}

\section{Harness Scoring}
\label{app:harness}

\paragraph{Export and scoring.}
For each changed file, the exporter writes a unified diff between the initial and final fixture and combines the diffs into one SWE-bench prediction per instance. Predictions are scored with the official harness (version 3.0.17) against the SWE-bench test split, which contains the Lite instances, and patches are scored as exported, without rewriting their context. Four issues seed two scenarios each, so their second predictions are scored in a separate pass. The earlier sweep's Opus 4.7 Lite predictions were scored against \texttt{princeton-nlp/SWE-bench\_Lite}. Harness jobs use one worker, instance-level image caching, no image cleanup, and a watchdog that terminates stalled jobs and reruns only the unfinished instances.

\paragraph{Optional judge.}
Some qualities resist predicates, for example whether a reviewer engaged with an owner's risk note. For these, \bench builds bounded evidence packets (scenario metadata, the deterministic scorecard, a changed-file summary, numbered trace events, and the owner-resource snippets the run accessed) for an LLM judge that rates constraint adherence and collaboration and must cite event ids. Outputs that cite nonexistent events are rejected.

\paragraph{Gold gate.}
Before scoring model patches, we send reference patches for 20 sampled instances (8 from SWE-bench Lite) through the same harness configuration; 15 of the 16 gold predictions that completed were resolved.

\section{Scoring from Trajectories}
\label{app:replay}

\paragraph{Replay.}
Each frontier trajectory lists the scenario, the agents with their tool sets, and every event: messages, tool calls, and the tools' observations. We assign each agent its role from its tool set and replay the observations in order, as the runtime applies them. Successful writes extend the changed files, owner-resource reads add the evidence keys that the tools return, and reviews, grants, test records, and the submission set the corresponding fields; refused calls change nothing. The trajectories replace file paths with tokens, which we map back to fixture paths by matching the content of reads and writes against the bundles. We then apply the predicates of Section~\ref{sec:evaluation} and the action checks of each scenario's contract. Because the bundles carry no canonical arguments, an action check passes when the agent in that role made at least one successful call of the tool, which is also how the pipeline's scorer treats them.

\paragraph{Owner evidence in cross-component bundles.}
The owners' notes and tables name every required file in 10 of the 50 cross-component bundles, some of them in 5, and only the first required file in 35. The opening message always lists all required files.

\section{Run Accounting}
\label{app:accounting}

Table~\ref{tab:lite-repo} breaks down the Lite harness results of Opus 4.7 by repository. Table~\ref{tab:accounting} lists, for each model of the earlier sweep, the valid runs, the model-attributable failures by type, and the runs that produced a scorecard, and Table~\ref{tab:archetype-all} gives its workflow and constraint success by archetype. Table~\ref{tab:frontier-arch} gives the same breakdown with \TS for the frontier sweep, in which every agent has a valid outcome on all 250 scenarios.

\begin{table}[!ht]
\centering
\small
\begin{tabular}{@{}lrr@{}}
\toprule
Repository & Resolved & Scenarios \\
\midrule
Django & 11 & 17 \\
SymPy & 7 & 12 \\
scikit-learn & 4 & 6 \\
Matplotlib & 3 & 4 \\
Astropy & 1 & 4 \\
xarray & 1 & 1 \\
pytest & 0 & 2 \\
Sphinx & 0 & 2 \\
Requests & 0 & 1 \\
Pylint & 0 & 1 \\
\midrule
Total & 27 & 50 \\
\bottomrule
\end{tabular}
\caption{Harness-resolved Opus 4.7 patches on the SWE-bench Lite scenarios of the earlier sweep, by repository.}
\label{tab:lite-repo}
\end{table}

\begin{table}[!ht]
\centering
\small
\begin{tabular}{@{}lrrrrrrr@{}}
\toprule
 & & \multicolumn{5}{c}{Model-attributable failures} & \\
\cmidrule(lr){3-7}
Model & Valid & Unparsable & Empty & Misrouted & Tool & Turns & Scored \\
\midrule
Claude Opus 4.7 & 250 & 0 & 0 & 0 & 0 & 0 & 250 \\
Claude Sonnet 4.5 & 220 & 7 & 0 & 0 & 0 & 0 & 213 \\
Qwen3-32B & 250 & 15 & 20 & 13 & 6 & 0 & 196 \\
\bottomrule
\end{tabular}
\caption{Run accounting of the earlier sweep. Model-attributable failures count as failures: unparsable replies, empty replies, misrouted messages, calls to a tool outside the agent's role (\emph{Tool}), and exhausted turn budgets (\emph{Turns}). \emph{Scored} runs produced a scorecard; runs that exhaust the turn budget can still be scored.}
\label{tab:accounting}
\end{table}

\begin{table}[!ht]
\centering
\footnotesize
\setlength{\tabcolsep}{4.5pt}
\begin{tabular}{@{}lccccc@{}}
\toprule
Model & Code review & QA acceptance & Security & Release gate & Cross-component \\
\midrule
Claude Opus 4.7 & 100.0 / 100.0 (50) & 98.0 / 100.0 (50) & 100.0 / 100.0 (50) & 88.0 / 90.0 (50) & 38.0 / 38.0 (50) \\
Claude Sonnet 4.5 & 93.8 / 93.8 (48) & 92.0 / 96.0 (25) & 78.0 / 78.0 (50) & 76.0 / 76.0 (50) & 38.3 / 38.3 (47) \\
Qwen3-32B & 8.0 / 52.0 (50) & 0.0 / 60.0 (50) & 8.0 / 34.0 (50) & 4.0 / 18.0 (50) & 2.0 / 24.0 (50) \\
\bottomrule
\end{tabular}
\caption{Workflow and constraint success (\WS / \CS, \%) by archetype in the earlier sweep, with the number of valid runs in parentheses.}
\label{tab:archetype-all}
\end{table}

\end{document}